\documentclass[letterpaper, 10 pt, conference]{ieeeconf}  

\let\labelindent\relax

\IEEEoverridecommandlockouts                              

\title{\LARGE \bf
GAMMA: Graspability-Aware Mobile MAnipulation \\
Policy Learning based on Online Grasping Pose Fusion
}

\author{Jiazhao Zhang$^{1,2,*}$, Nandiraju Gireesh$^{3,*}$, Jilong Wang$^2$, Xiaomeng Fang$^2$, \\ Chaoyi Xu$^3$, Weiguang Chen$^2$, Liu Dai$^4$, and He Wang$^{1,2,3,\dagger}$
\thanks{*Joint first authors}
\thanks{$^{1}$CFCS, School of Computer Science, Peking University. \quad
$^{2}$Beijing Academy of Artificial Intelligence \quad
$^{3}$Galbot \quad
$^{4}$Tongji University}%
}

\usepackage{mathtools}
\usepackage{amsmath}

\usepackage{amssymb}
\usepackage{amsfonts}
\usepackage{amsopn}
\usepackage{graphicx} 
\usepackage{textcomp}
\usepackage{xfrac}
\usepackage{bbm}
\usepackage{overpic}
\usepackage{subfig}
\usepackage{wrapfig}
\usepackage[ruled,vlined,linesnumbered]{algorithm2e}
\SetAlFnt{\small}
\SetAlCapFnt{\small}
\SetAlCapNameFnt{\small}
\SetAlCapHSkip{0pt}

\usepackage{multirow}

\usepackage{booktabs}
\usepackage{footnote}
\usepackage{paralist}
\usepackage{enumitem}
\usepackage{autobreak}
\usepackage{hyperref}
\usepackage{cleveref}

\usepackage{color}
\definecolor{green}{rgb}{0, 0.4, 0}
\definecolor{orange}{rgb}{0.8, 0.6, 0.2}
\definecolor{red}{rgb}{1.0, 0.0, 0.0}
\definecolor{teal}{rgb}{0.0, 0.4, 0.4}
\definecolor{purple}{rgb}{0.65,0,0.65}
\definecolor{saffron}{rgb}{0.95,0.75,0.2}
\definecolor{turquoise}{rgb}{0.0,0.5,0.5}
\definecolor{brown}{rgb}{0.5, 0.16, 0.16}

\usepackage{overpic}
\usepackage{currfile} 

\newlength\savedwidth

\definecolor{lightgray}{rgb}{0.6, 0.6, 0.6}

\newcommand{\addcite}[1]{{\textcolor{red}{[cite]}}}

\definecolor{revisedcolor}{RGB}{100,0,200}

\usepackage[normalem]{ulem}

\newcommand{\hidecomment}[1]{}

\newcommand{\cS}{\mathcal{S}}

\newcommand{\cG}{\mathcal{G}}

\newcommand{\cA}{\mathcal{A}}

\usepackage{xspace}

\usepackage[table]{xcolor}
\begin{document}

\maketitle
\thispagestyle{empty}
\pagestyle{empty}


\begin{abstract}

Mobile manipulation constitutes a fundamental task for robotic assistants and garners significant attention within the robotics community. A critical challenge inherent in mobile manipulation is the effective observation of the target while approaching it for grasping. In this work, we propose a graspability-aware mobile manipulation approach powered by an online grasping pose fusion framework that enables a temporally consistent grasping observation. Specifically, the predicted grasping poses are online organized to eliminate the redundant, outlier grasping poses, which can be encoded as a grasping pose observation state for reinforcement learning. Moreover, on-the-fly fusing the grasping poses enables a direct assessment of graspability, encompassing both the quantity and quality of grasping poses. 
This assessment can subsequently serve as an observe-to-grasp reward, motivating the agent to prioritize actions that yield detailed observations while approaching the target object for grasping.
Through extensive experiments conducted on the Habitat and Isaac Gym simulators, we find that our method attains a good balance between observation and manipulation, yielding high performance under various grasping metrics.
Furthermore, we discover that the incorporation of temporal information from grasping poses aids in mitigating the sim-to-real gap, leading to robust performance in challenging real-world experiments. Project page: \url{https://pku-epic.github.io/GAMMA/}


\end{abstract}

%


\section{INTRODUCTION}

Autonomous mobile manipulation has been an essential research area in robotics~\cite{brock2016mobility, hebert2015mobile}, leading to diverse applications, \textit{e.g.}, manufacturing, warehousing, construction, and household assistance~\cite{watkins2022mobile, kalashnikov2018qt, mahler2019learning,vivaldini2010robotic}. 
For research on mobile manipulation, a challenging but popular task setting is to require
the agent to actively observe and explore an unseen environment with the goal of manipulating a target object. Originated from the unseen nature of the environment, the agent can't directly plan a trajectory to reach and grasp the object. Instead, its has to rely on online observations and scene priors to make a success, posing many research questions to the area. 

Many existing works tackle this problem via combining 3D reconstruction and scene geometry analysis with motion planning~\cite{sun2022motion, chen2023hierarchical, patki2020language}. However, such approaches usually suffer from huge computational costs for modeling the scene geometry and entail complicated heuristic designs tailored to specific robots. Recently, reinforcement learning (RL) based approaches~\cite{yokoyama2023adaptive, jauhri2022robot, wang2020learning, sun2022fully} have gained more attention due to their simplicity and efficiency. 
Exemplar works~\cite{jauhri2022robot, watkins2022mobile} use visibility and reachability as scene priors, which can drive the agent to observe and approach the target object.
They propose to learn and use them in policy's input states as well as reward, which has significantly improved the policy performance.
%

%
%

\begin{figure}[t]
\centering
\begin{overpic}
[width=0.95\linewidth]
{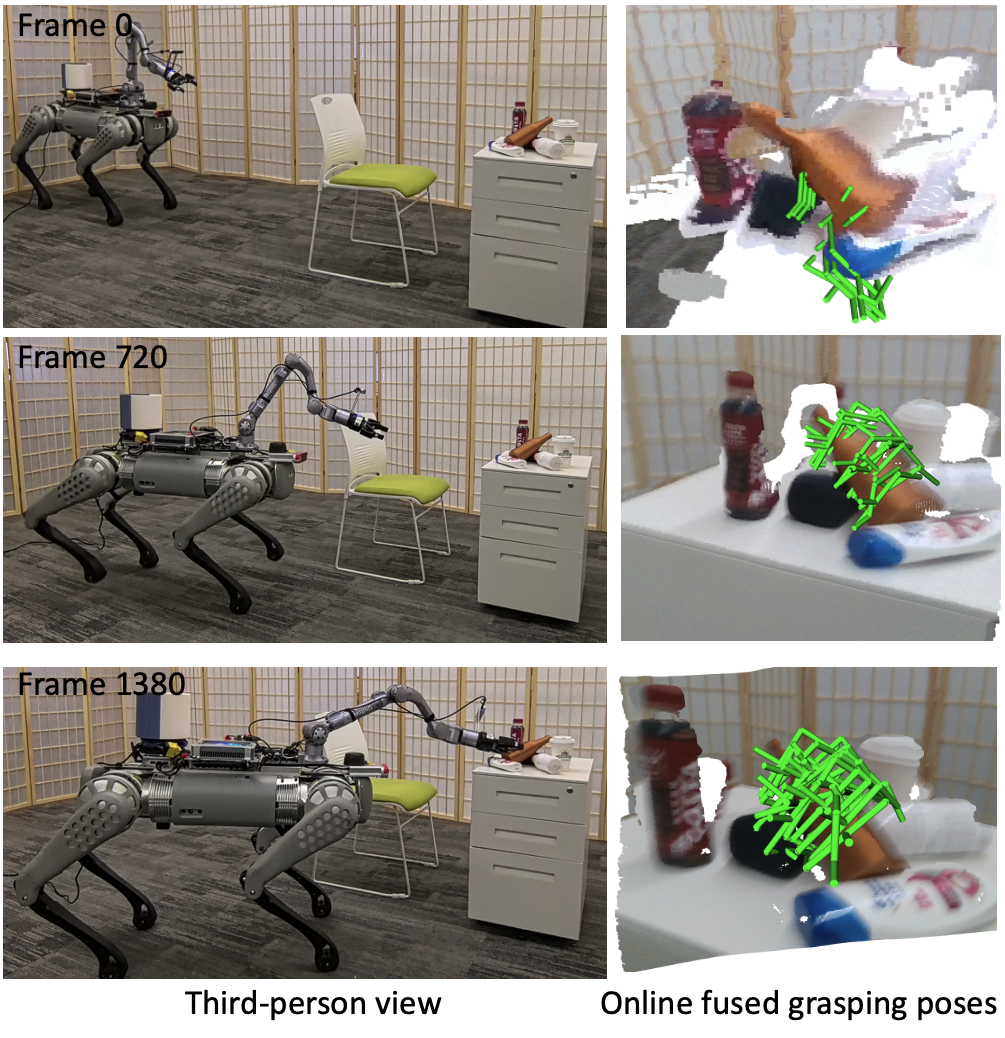}
\end{overpic}
\caption{
We present a graspability-aware mobile manipulation approach powered by an online grasping pose fusion framework that enables a temporally consistent grasping observation and efficient grasping.
}
\vspace{-6mm}
\label{fig:grasping_fusion}
\end{figure}

In this work, we focus on an RL-based approach and propose a novel scene prior,
\textit{graspability}, to advance mobile manipulation policy learning. We define graspability as a complete set of valid grasping poses of the target object. Compared with reachability and visibility, which provide circuitous guidance to the agent for graspings, graspability offers more direct and informative guidance for effective grasping guidance.



Note that graspability contains the full information of the valid target object graspings, online estimating graspability in an unseen environment is highly non-trivial,
\textit{e.g.}, online observations during mobile manipulation often include many occlusions as well as large overlaps, leading to noises and redundancy in grasping pose predictions~\cite{Wang2021GraspnessDI}.
We thus propose an online grasping pose fusion module, which dynamically fuses the redundant grasping poses and removes the outlier poses. This fusion process yields high-quality graspability estimation that achieves high precision and recall of valid grasping poses.

To facilitate agent learning, we propose the following two ways to fully utilize the estimated graspabiltity. First, we propose to encode graspabiltity into states and use it in the policy input, endowing the agent with the awareness of grasping goals. We find that our graspability-aware agent can thus learn to move its base and arm more intelligently.
Second, we propose to use the number of grasps and the distance-to-grasp information in graspability as RL reward, encouraging the agent to gain more observations of valid grasping poses. We also introduce a weight schedule that combines these two rewards to balance the observation goal and the grasping goal. This reward motivates the agent to prioritize extensive observations in the initial stages, subsequently shifting its focus to grasp the target object.

Through extensive experiments on two mainstream simulators, Habitat~\cite{Szot2021Habitat2T} and Isaac Gym~\cite{makoviychuk2021isaac}, which include a diverse range of environments and objects, we demonstrate that our method outperforms mainstream methods on both abstract grasping metric and realistic grasping pose metric. Moreover, real-world evaluations of our approach further showcase the robustness and effectiveness of our methodology.


In summary, the contributions of our work include:

\begin{itemize}

\item We propose an online grasping fusion module to fuse predicted grasping poses to obtain temporally consistent grasping poses for erasability observation.

\item We design an observe-to-grasp reward to effectively encourage agents to execute actions that balance both observation and grasping.

\item We present a graspability-aware mobile manipulation RL system, achieving robust performance on both simulators and real-world environments.

\end{itemize}

\begin{figure*}[ht]
\centering
\begin{overpic}
[width=0.95\linewidth]
{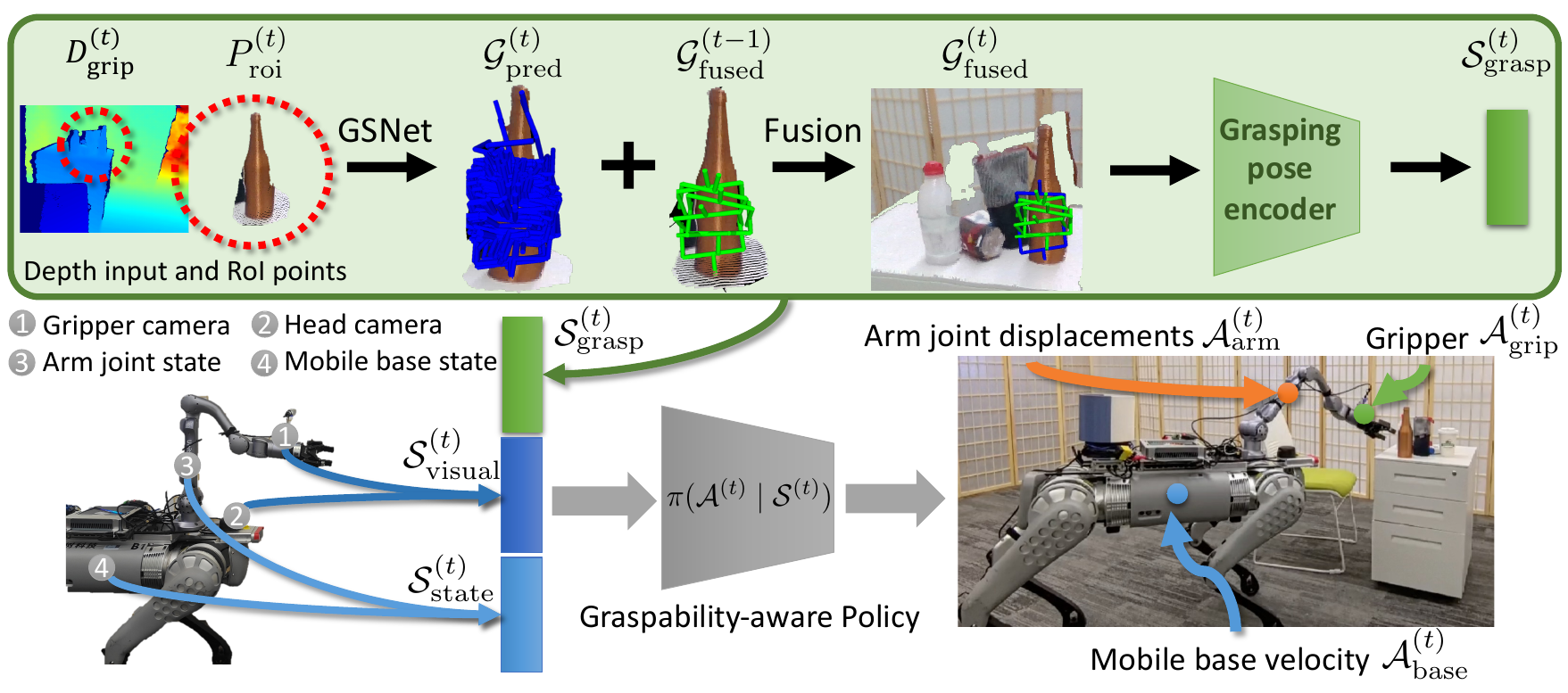}
\end{overpic}
\caption{Method overview. Our method processes the gripper depth map $D_\text{grip}^{(t)}$ to region-of-interest point cloud $P_\text{roi}^{(t)}$, which then be sent to GSNet for predicting grasping poses $\cG_\text{pred}^{(t)}$. The $\cG_\text{pred}^{(t)}$ will then be integrated into the so far fused grasping poses $\cG_\text{fused}^{(t-1)}$ to obtain $\cG_\text{fused}^{(t)}$. The $\cG_\text{fused}^{(t)}$ will be encoded as $\cS_\text{grasp}^{(t)}$, along with  $\cS_\text{visual}^{(t)}$ and $\cS_\text{state}^{(t)}$ for learning $\pi(\cA^{(t)} \mid \cS^{(t)})$.}

\vspace{-0.5cm}
\label{fig:pipeline}
\end{figure*}

\section{Related work}
\label{sec:relatedworks}

\textbf{Traditional mobile manipulation methods.} 
For decades, the field of robotics has experienced significant growth in the advancement of mobile manipulation methods~\cite{brock2016mobility, hebert2015mobile}. 
Traditional researches~\cite{heins2021mobile, sun2022motion, garrett2020replan} leverage scene analysis and motion planning, aiming to devise strategies for efficient task execution. However, these approaches assume access to explicit secure information regarding the environments, such as detailed maps with obstacle locations~\cite{heins2021mobile, patki2020language,zheng2019active}, precise object coordinates~\cite{garrett2020replan, sun2022motion, chen2023hierarchical, zhang20233d}.


\textbf{Learning-based mobile manipulation methods.}
Mobile manipulation agents are trained to possess the capability to observe and interact within various scenes. One of the primary capabilities of these agents is to observe the target~\cite{he2022visibility, xu20163d}, achieved by encouraging the agent to obtain multiple observations of the target object. 
Another crucial capability is to maneuver its arm to approach the target object, often referred to as reachability~\cite{yokoyama2023adaptive, jauhri2022robot, dai2023graspnerf}. Regarding graspability, advanced methods do not rely on predicted grasping poses~\cite{Bajracharya2023DemonstratingMM} or object pose estimation~\cite{yamazaki2022approaching, chowdhury2023neural}, as these approaches lack temporal perception of graspability. 
In this paper, we introduce an online grasping pose fusion module to fuse the predicted grasping poses for encoding graspability states, enabling our method to be graspability-aware and showcasing enhanced performance.


\section{Problem Statement and Method Overview}
\label{sec:overview}

\textbf{Mobile manipulation task.} Given a target object location $p_{\text{goal}}$, the robot is tasked with navigating through an unknown environment to effectively approach and grasp the target object. We follow the mainstream setup presented in \cite{jauhri2022robot, yokoyama2023adaptive}. The robot is equipped with a mobile base, an arm, and a parallel gripper. Two RGB-D cameras are mounted: one to the head of mobile base ($D_{\text{head}}, I_{\text{head}} $) and the other to the gripper ($D_{\text{grip}}$,$I_{\text{grip}}$), where $d$ and $c$ represent the depth image and color image, respectively. The robot utilizes a 3-DoF configuration for its mobile base in SE(3), coupled with an $(x+1)$-DoF arm. In detail, $x = 6$ for the Spot arm and the Unitree Z1 arm, and $x = 7$ for the Fetch robots, further augmented by a 1-DoF gripper for object grasping. 

\textbf{Overview.} Figure~\ref{fig:pipeline} provides an overview of our proposed graspability-aware mobile manipulation approach. To facilitate graspability, our method processes the depth image $I^d_{\text{grip}}$ and leverages an off-the-shelf grasping module, GSNet~\cite{Wang2021GraspnessDI}, to predict grasping poses (Section~\ref{sec:grasp}). These predicted grasping poses are then fused online (Section~\ref{sec:ogfm}) and encoded as the graspability state $\cS_{\text{grasp}}$. Subsequently, our method can learn the graspability-aware mobile manipulation policy $\pi(\cA_{\text{base}},\cA_{\text{arm}},\cA_{\text{grip}}|\cS_{\text{grasp}}, \cS_\text{visual},\cS_{\text{state}})$ through reinforcement learning, incorporating visual information $\cS_\text{visual}$ and state information $\cS_{\text{state}}$ (Section~\ref{sec:grasp_obs}).

In our method, the policy generates a 3-DoF SE(3) velocity for mobile base control $\cA_{\text{base}}$, a 6-DoF residual adjustment for current arm joints $\cA_{\text{arm}}$, and a 1-DoF switch to control the gripper $\cA_{\text{grip}}$.
During the RL training process, a composite observe-to-grasp reward is employed, incorporating both the grasping observation reward, $g_{\text{go}}$, and the gripper-to-grasping poses reward, $g_{\text{gg}}$. This reward system motivates the robot to prioritize actions based on meticulous observations, guiding it towards more optimal grasping poses (Section~\ref{sec:rl}). For ease of description, the notations used in this paper default to the world coordinate system.

\section{Graspability Estimation}
To obtain accurate and complete graspability of an object, we propose to predicts grasping poses at each timesteps (see \ref{sec:grasp}) and online fuses them together while eliminating invalid ones (see \ref{sec:ogfm}). 


\subsection{Grasping pose prediction}
\label{sec:grasp}

During mobile manipulation, our graspability-aware agent constantly captures observations from RGB-D cameras and performs online predictions of grasping poses. 
At each time step $t$, the agent moves according to the policy, and then online obtains new observations of the scene.
To obtain grasping poses $\cG^{(t)}$ based on the agent's observation, we leverage GSNet~\cite{Wang2021GraspnessDI}, which is trained on a billion-scale real-world dataset~\cite{Fang2020GraspNet1BillionAL} and demonstrates robust performance in novel scenes~\cite{Gou2022UnseenO6, Fang2022AnyGraspRA}. Given the target object's location $p_\text{goal}^{(t)}$, we can extract a sphere-shaped region-of-interest (RoI) point cloud $P_\text{roi}^{(t)}$ from the 3D points $P_\text{grip}^{(t)}$. These points are obtained by back-projecting the depth map $D_\text{grip}^{(t)}$ and then be transformed to the world coordinate system. Therefore, the $P_\text{roi}^{(t)}$ can be formulated as follows: 
\begin{equation}
    P_\text{roi}^{(t)} = \{P_\text{grip}^{(t)}\in \mathbb{R}^3 \mid \|P_\text{grip}^{(t)} - p_{\text{goal}}\|_2 < \tau \},
\end{equation}
where $\tau$ represents the maximum distance from the target object's location. Empirically, we set $\tau = 10$ cm. Then we can utilize GSNet $G(\cdot)$ to predict the grasping poses as follows:
\begin{equation}
    \cG_\text{pred}^{(t)} =  G(P_\text{roi}^{(t)}) = \{q_i^{(t)}, p_i^{(t)}, s_i^{(t)}\}_{i=1:n},
\end{equation}
$q$, $p$, and $s$ correspond to the orientation (represented by quaternion), position, and score of the predicted poses, respectively. Additionally, $n$ denotes the number of grasping poses, which may be zero if the quality of $P_\text{grip}^{(t)}$ is low.



\subsection{Online Grasping Fusion Module.}
\label{sec:ogfm}
Note that these predicted grasping poses $\cG_\text{pred}^{(t)}$ from each timestep can be noisy due to occlusions and overlapping significantly with previous ones.
As a result, directly combining all predicted grasping poses $\cG_\text{pred}^{(0)}\cup... \cup\cG_\text{pred}^{(t)}$ to obtain graspability may lead to many errors and a high degree of redundancy, posing further challenges to policy learning. To address this, we introduce an online grasping fusion module, which is designed to maintain temporally consistent grasping pose observations. An illustration of the fusion module is presented in Figure~\ref{fig:grasping_fusion}.


To store and track the predicted grasping poses, our fusion module first partitions the 3D space into a uniform grid in the center of the target object location $p_{goal}$. Here, we utilize a $64\times64\times64$ cube grid ($3$cm voxel). For efficient memory usage, we use an off-the-shelf indexing table algorithm~\cite{Zhang2020FusionAwarePC} to dynamically allocate the voxels. For each grasping pose $\cG_\text{fused}^{(t)} =  \{o_i^{(t)}, p_i^{(t)}, s_i^{(t)}\}_{i=1:n}$, its corresponding voxel $v$ can be found by $v^{min}_{x/y/z} < p^{(t)}_{i,x/y/z}<v^{max}_{x/y/z}$.
where $x/y/z$ represents the comparisons across the three axes. With this approach, each voxel stores a collection of grasping poses, allowing for easy identification of neighboring grasping poses within a specified 3D range.

However, these grasping poses are unorganized (lacking spatial information to one another) and redundant, requiring further refinement by merging grasping poses. Recognizing that grasping poses within a voxel are more sensitive to orientation than to translation~\cite{Fang2022AnyGraspRA}, our method retains only those grasping poses that exhibit a considerable angular difference compared to other existing poses within the same voxel.

Specifically, our method iteratively calculates the angle between new grasping poses $\cG_\text{pred}^{(t)}$ and fused grasping poses $\cG_\text{fused}^{(t-1)}$ belonging to the same voxel. If the angle exceeds $\tau_\text{angle}$, the new grasping pose will be added to the voxel grasping set. Note that, we empirically set the $\tau_{\text{angle}}=\pi/4$ through experiments. Moreover, grasping poses are mainly located within voxels that are close to the target objects, making the grasping pose query highly efficient.

Given such a fused grasping grid, we can efficiently traverse saved grasping poses to identify the $\{q_\text{fused}^{(t-1)}, p_\text{fused}^{(t-1)}, s_\text{fused}^{(t-1)}\}$ with the smallest angle differences. If the angle is less than $\tau_\text{angle}$, we utilize a weighted average with the weight determined by the score of grasping poses, represented as $w=s_{\text{new}}/(s_{\text{fused}} + s_{\text{new}})$:
\begin{equation}
\begin{split}
        p_\text{fused}^{(t)} &= (1-w) p_{\text{fused}}^{(t-1)} + w p_\text{pred}^{(t)},\\
        q_\text{fused}^{(t)} &= \frac{\sin((1-w)\theta)}{\sin(\theta)} q_\text{fused}^{(t-1)} + \frac{\sin(w\theta)}{\sin(\theta)} q_\text{pred}^{(t)},\\
        s_\text{fused}^{(t)} &= s_{\text{fused}}^{(t-1)} + s_\text{pred}^{(t)},
\end{split}
\end{equation}
where the $q_\text{fused}$ is renormalized to $1$ to satisfy the quaternion constraint. 
Note that, the updated orientation $q_{\text{fused}}$ may break the angle distance constraints, therefore new fused grasping pose will then be compared with other grasping poses within the same voxel until all the grasping poses satisfy the angle distance threshold. We find that the recursive grasping pose fusion is a rare occurrence due to the sparse distribution of grasping poses (typically containing approximately $4$ grasping poses) in our implementation. This fusion operation leads to complete and accurate fused grasping pose results.  

\begin{figure}[t]
\centering
\begin{overpic}
[width=\linewidth]
{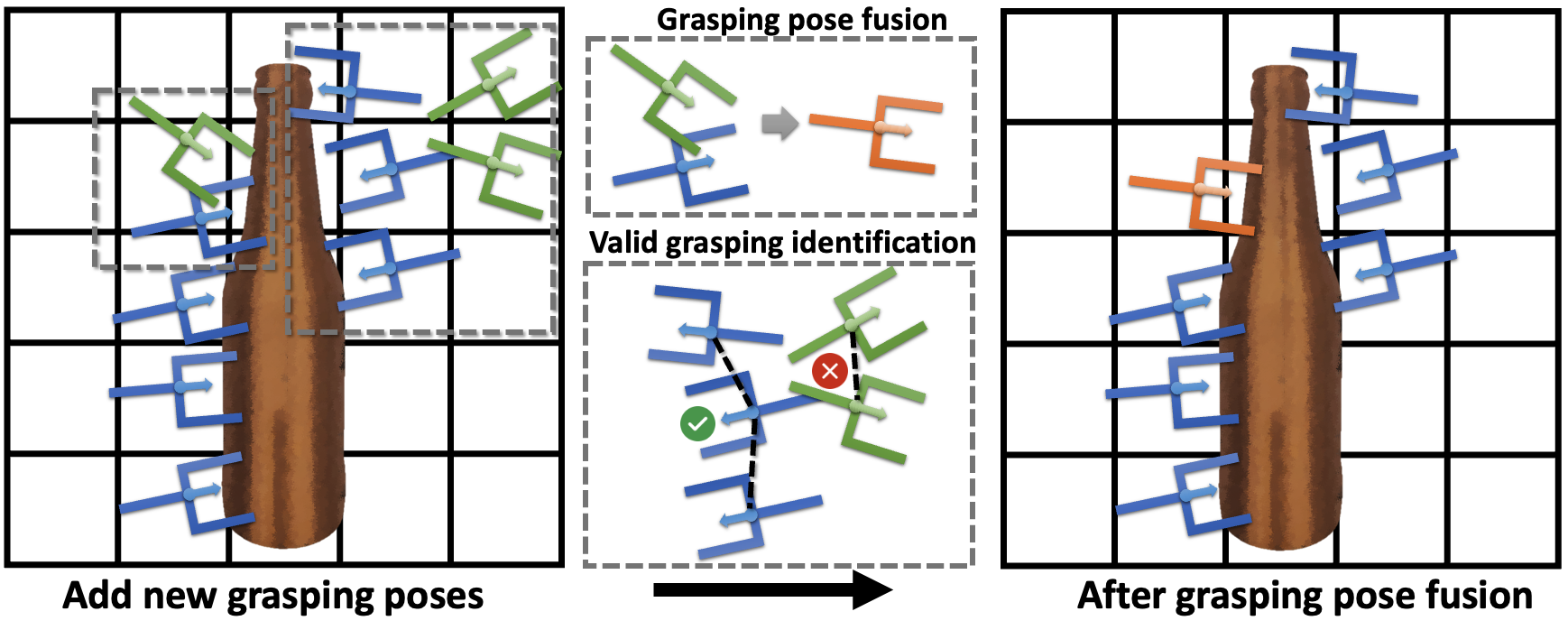}
\end{overpic}
\caption{
An illustration of grasping pose fusion and valid grasping pose identification. New added grasps and previous fused grasping poses are indicated in \textcolor{green}{green} and \textcolor{blue}{blue}, respectively.
}
\vspace{-6mm}
\label{fig:grasping_fusion}
\end{figure}

\textbf{Valid grasping pose identification.} Due to occlusions arising from observational viewpoints, the prediction outcomes may include invalid grasping poses. These grasping poses could be distant from or oriented away from the target object, as depicted in Figure~\ref{fig:grasping_fusion}. To remove the outlier grasping poses, we follow the basic fact that the grasping pose should be densely and tangentially distributed along the target objects. Hence, we design a grasping pose consistency verification algorithm to evaluate the density of the grasping cluster. 


Specifically, our method connects neighboring grasping poses to form a 'grasping cluster' based on two criteria: 
(1) \textit{Distance}: The grasping poses should locate in adjacent voxels. 
(2) \textit{Orientation}: The angular difference between orientations should be less than $1.5\tau_{\text{angle}}$. 
Upon establishing these connections, clusters containing fewer grasping poses than $\tau_\text{count}$ are eliminated. These smaller clusters typically consist of outlier or error-prone poses. The resulting set preserves only the high-quality grasping poses for graspability observation.



\section{Graspability-aware policy learning.}

\subsection{Graspability states }  
\label{sec:grasp_obs}

Taking the output grasping poses (a.k.a. graspablity) from the online grasping fusion module, we propose to encode them as a part of the state for RL.

Note that this encoding is highly non-trivial because the grasping poses are unordered high-dimensional vectors. As a part of the pose, the quaternion $q$ that represents grasping pose orientation is discontinuous among SO(3) manifold~\cite{Zhou2018OnTC,chen2022projective}, further creating difficulties.
To tackle these challenges, we first convert quaternions into continuous 6D rotation representation $F(\cdot)$ ~\cite{Zhou2018OnTC}; and then leverage an order-invariant neural network PointNet~\cite{Qi2016PointNetDL} (composed by three MLP layers $M$ and a maxpooling layer) for state encoding:
\begin{equation}
\label{equ:grasp}
    S_{\text{grasp}} = maxpooling\{M(p_\text{fused},F(q_\text{fused}),s_\text{fused})\},
\end{equation} 
A corner case we need to handle: at the beginning of mobile manipulation, the camera hasn't observed the target object yet, there is no grasping pose available .
In this case, we need our graspability-aware approach to degenerate into being reachability-aware, directing the agent towards the target object's location.

We thus leverage the same encoding method of graspability states (Equation~\ref{equ:grasp}), and substitute the grasping pose with the target object location $p_{\text{goal}}$, uniform sampled orientation $q_{\text{sample}}$ within SO(3), and a constant low score $s_{\text{reach}}$ (set to $0.1$) to form the reachability observation $ S_{\text{reach}} \approx S_{\text{grasp}} $:
\begin{equation}
\label{eqy:reach}
    S_{\text{reach}}= maxpooling\{M(p_{\text{goal}},F(q^{(k)}_{\text{sample}}),s_{\text{reach}})\}_{k=1:K}.
\end{equation}
The uniformly sampled orientation $q^{(k)}_{\text{sample}}$ ($K=128$) encourages the arm to reach the target object from any direction until the online grasping pose fusion module supplies valid grasping poses.
Besides the graspability state, we also encoded the visual information $\cS_\text{visual}$ from both the front camera and gripper cameras and $\cS_\text{state}$ from joints state encoding. These follow the same methodology as described in~\cite{Szot2021Habitat2T}. Consequently, the graspability-aware policy can be expressed as $\pi(\cA_{\text{base}},\cA_{\text{arm}},\cA_{\text{grip}}|\cS_\text{grasp}, \cS_\text{visual}, \cS_\text{state})$. 


\subsection{Observe-to-grasp reward for RL training.} 

\label{sec:rl}

With the graspability observation, the agent is required to make a balance between observing and grasping during mobile manipulation. To this end, we design an observe-to-grasp reward mechanism, consisting of a grasping observation reward $r_{\text{go}}$ and a gripper-to-grasping pose reward $r_{\text{gg}}$:
\begin{equation}
\begin{split}
        r_{\text{go}}^{(t)} &= \sum s^{(t)}_{\text{fused}} - \sum s^{(t-1)}_\text{fused}, \\
         r_{\text{gg}}^{(t)} &= D_{\text{gg}}^{(t-1)} - D_{\text{gg}}^{(t)}, 
\end{split}
\end{equation}
where $s^{(t)}_{\text{fused}} \in \cG^{(t)}_{\text{fused}}$, and $D_{\text{gg}}$ is the gripper to grasping pose evaluation function. For $r_{\text{go}}$, we leverage the online grasping fusion module, directly assessing the enhancement of graspability observation. And for the gripper-to-grasping reward $r_{\text{gg}}$, the gripper is encouraged to approach to fused grasping pose with a high score:
\begin{equation}
\label{equ:dgg}
\begin{split}
    D_{\text{gg}}&= \min\{\ e^{-s_\text{fused}} ( \beta_1 \|  p_{\text{grip}} - p_\text{fused} \|_2 + \beta_2 \theta(q_{\text{grip}},q_\text{fused})) \},
\end{split}
\end{equation}
where  $\theta(\cdot)$ compute the interval angle between two rotations (radius). And $\beta_1$ and $\beta_2$ are set to $0.3$ and $0.2$, respectively.
Finally, we can formulate our reward as: 
\begin{equation}
\begin{split}
    r_{\text{o2g}}^{(t)} & = (1-\sigma) r_{\text{go}}^{(t)} + \sigma r_{\text{gg}}^{(t)},\\
    \sigma & = \frac{1}{1+e^{0.5-t/t_\text{max}}},
\end{split}
\end{equation}
and the $\sigma$ is a logistic sigmoid function related to the execution steps. This approach ensures that the observe-to-grasp reward initially encourages the agent to observe, and as more steps are taken, gradually shifts to promoting grasping actions. Such a reward is dense and adaptive, guiding the agent's learning more effectively.
In addition to the observe-to-grasp reward, we also leverage a sparse success reward $r_{\text{success}}=10$ ($D_{\text{ee}}<15\text{cm}$), a slack penalty $r_{\text{slack}}=10^{-2}$ and a force penalty $r_{\text{force}}=10^{-4}$. These additional rewards enhance the stability of the learning process.

\textbf{Implementation details.} We use Habitat 2.0 as our training simulator. Our method predicts sample actions every 20 steps and uses Proximal Policy Optimization (PPO)~\cite{Schulman2017ProximalPO} for agent training. Given the substantial overlap between consecutive frames for increased efficiency, our method predicts grasping poses every $10$ frames. 
We uniformly sample $128$ fused grasping poses (allowing repetition if the number of fused grasping poses is fewer than $128$) for $\cS_\text{grasp}$ (Equ.~\ref{equ:grasp}). In the absence of grasping poses, we resort to using $\cS_\text{reach}$ (Equ.~\ref{eqy:reach}). These parameters, including thresholds and size, are adopted through experiments and can be further improved with careful adjustments for specific scenes. Any parameters not detailed in this paper are adopted from~\cite{Szot2021Habitat2T}.





\begin{table*}[ht]\centering
\caption{
Quantitative comparison and ablation study on the Habitat simulator on both non-cluttered and cluttered episodes.
}
\scalebox{0.95}{
\setlength{\tabcolsep}{4mm}
\begin{tabular}{lcccccccc}
\hline
\multicolumn{1}{l|}{}                          & \multicolumn{4}{c|}{Fetch Robot}                                                                                        & \multicolumn{4}{c}{Unitree B1+Z1 arm}                                                                  \\ \cline{2-9} 
\multicolumn{1}{l|}{}                          & \multicolumn{2}{c|}{Non-cluttered env.}                  & \multicolumn{2}{c|}{Cluttered env.}                      & \multicolumn{2}{c|}{Non-cluttered env.}                  & \multicolumn{2}{c}{Cluttered env.}  \\ \cline{2-9} 
\multicolumn{1}{l|}{\multirow{-3}{*}{Methods}} & \multicolumn{1}{c|}{GazeSR} & \multicolumn{1}{c|}{GraspSR} & \multicolumn{1}{c|}{GazeSR} & \multicolumn{1}{c|}{GraspSR} & \multicolumn{1}{c|}{GazeSR} & \multicolumn{1}{c|}{GraspSR} & \multicolumn{1}{c|}{GazeSR} & GraspSR \\ \hline \hline
\rowcolor[HTML]{D4D4D4} 
HB\cite{Szot2021Habitat2T}                               & 45.8                         & -                          & 22.0                         & -                          & 57.0                         & -                          & 31.9                         & -     \\
M3\cite{Gu2022MultiskillMM}                             & 55.2                         & 49.4                          & 43.6                         & 39.5                          & -                         & -                          & -                         & -     \\
\rowcolor[HTML]{D4D4D4} 
ReachMM                                        & 36.5                         & 29.1                          & 18.2                         & 11.3                          & 39.1                         & 29.4                          & 21.5                         & 12.7     \\
GAMMA without fusion                                  & 10.3                         & 7.2                          & 8.1                         & 3.2                          & 10.9                         & 7.3                          & 8.1                         & 2.1     \\
\rowcolor[HTML]{D4D4D4} 
\textbf{GAMMA (Ours)}                                    & \textbf{67.3}                         & \textbf{62.4}                          & \textbf{60.7}                         & \textbf{57.1}                          & \textbf{71.0}                         & \textbf{69.2}                          & \textbf{66.3}                         & \textbf{64.5}     \\ \hline
\end{tabular}
}
\vspace{-3mm}
\label{tab:benchmark}
\end{table*} 

\section{Experiments}

\subsection{Experimental setup}
\label{sec:exp_set_up}

\begin{figure}[t]
\centering
\begin{overpic}
[width=0.85\linewidth]
{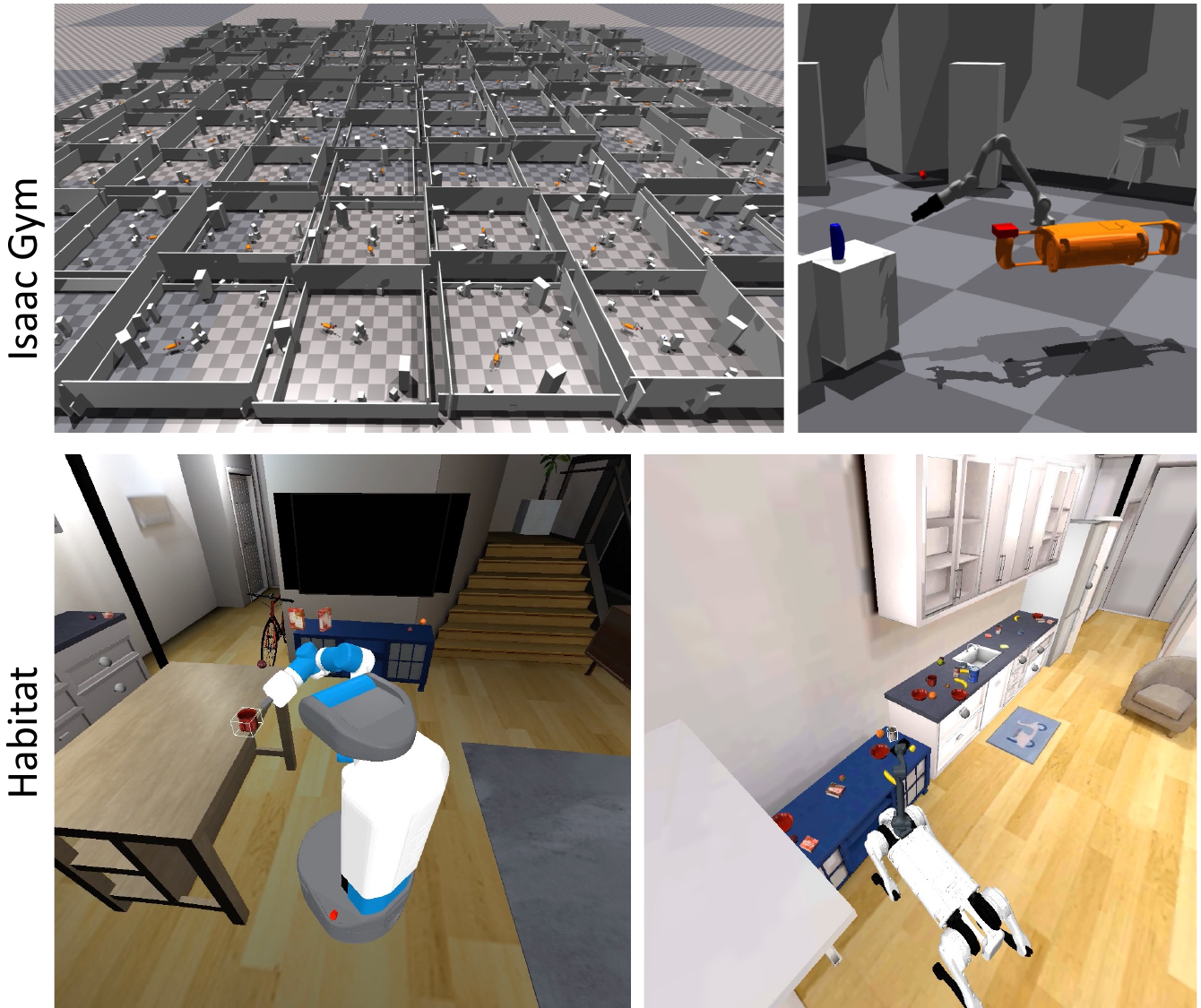}
\end{overpic}
\caption{
Simulation setup. First row: Unitree B1 + Z1 robot dog in Isaac Gym. Second row: Fetch Robot (left) and Unitree B1 + Z1 robot dog (right) in Habitat Simulator.
}
\vspace{-5mm}
\label{fig:exp_setup}
\end{figure}

\textbf{Synthetic environment setup.} We evaluate our method on the Habitat 2.0 simulator~\cite{Szot2021Habitat2T} and Isaac Gym~\cite{makoviychuk2021isaac}. The Habitat 2.0 features photo-realistic reconstructions of apartment scenes from ReplicaCAD~\cite{Szot2021Habitat2T}, stuffed by objects from YCB dataset~\cite{alli2015TheYO}.
We use both two datasets on Habitat, including 1000 Habitat episodes and a self-build 1,000 challenging episode dataset with cluttered object layouts (approximately 10 objects on each receptacle). For Isaac Gym, we create a mobile manipulation environment similar to ~\cite{jauhri2022robot}.
\textit{The episode data will be released to the public.}

\textbf{Real-world environment setup.}  We deploy the B1+Z1 robot in the real world to grasp the target object from a cluttered receptacle (with 4-6 objects) while performing obstacle avodiance. In detail, We mount an Azure Kinect DK on the head of the B1 robot dog and a Realsense D415 on the gripper of the Z1 arm. During the experiments, we make use of ORB-SLAM3~\cite{campos2021orb} to obtain 6D grasping poses based on the gripper camera observations. Note that the extrinsic parameters between the cameras and the arm, as well as the robot base, are pre-calibrated. 


\textbf{Baselines.} 
Given the intricacy of the task, ensuring a fair comparison among all mainstream methods is a formidable challenge. Hence, we focus on comparing with other methods that are most pertinent to our approach and have been assessed within the same simulator environment. Specifically, we consider:
\begin{enumerate}
    \item Multi-skill Mobile Manipulation \textbf{(M3)}~\cite{Gu2022MultiskillMM}: A modular method which incorporates mobility for enhanced flexibility in object interactions.
    \item Habitat-Baselines \textbf{(HB)}: A standard baseline method provided by Habitat 2.0.
    \item Reachability-aware policy \textbf{(ReachMM)}: An approach that leverages the reachability encoded state $\cS_\text{reach}$ as defined in Equ.~\ref{eqy:reach}.
    \item Non-fusion graspability-aware policy \textbf{(GAMMA without fusion)}: A method which doesn't make use of our proposed OGFM module and predicts grasp poses for every frame.
\end{enumerate}

\textbf{Metrics.} We measure the episode when the grasp action is called. In simulator, we consider two following metrics: (1) \textit{Gaze Success Rate} (GazeSR), an episode is deemed successful if the distance between the arm camera position and the target object position is within the $15$ cm and the angle between the camera ray and the object-to-camera ray is less than $10^\circ$. (2) \textit{Grasp Success Rate} (GraspSR), Success is achieved if the gripper's pose closely matches any densely annotated grasping pose, with deviations less than $10$ cm in distance and $10^\circ$ in angle. 

\begin{table}[!t]\centering
\caption{
Comparisons on Isaac Gym simulator and real-world experiments under challenging conditions (with obstacles and cluttered objects).
}
\begin{tabular}{lccc}
\hline
\multicolumn{1}{l|}{}                          & \multicolumn{2}{c|}{Isaac Gym Simulator}                   & Real-world Env. \\ \cline{2-4} 
\multicolumn{1}{l|}{\multirow{-2}{*}{Methods}} & \multicolumn{1}{c|}{GazeSR} & \multicolumn{1}{c|}{GraspSR} & SR             \\ \hline \hline
GAMMA without fusion & 89.0\% & 29.6\% & 53.3\% \\
GAMMA (Ours)     & \textbf{96.6\%} & \textbf{86.6\%} & \textbf{73.3\%} \\ \hline
\end{tabular}
\vspace{-0mm}
\label{tab:comp_real}
\end{table}

\begin{figure}[t]
\centering
\begin{overpic}
[width=0.95\linewidth]
{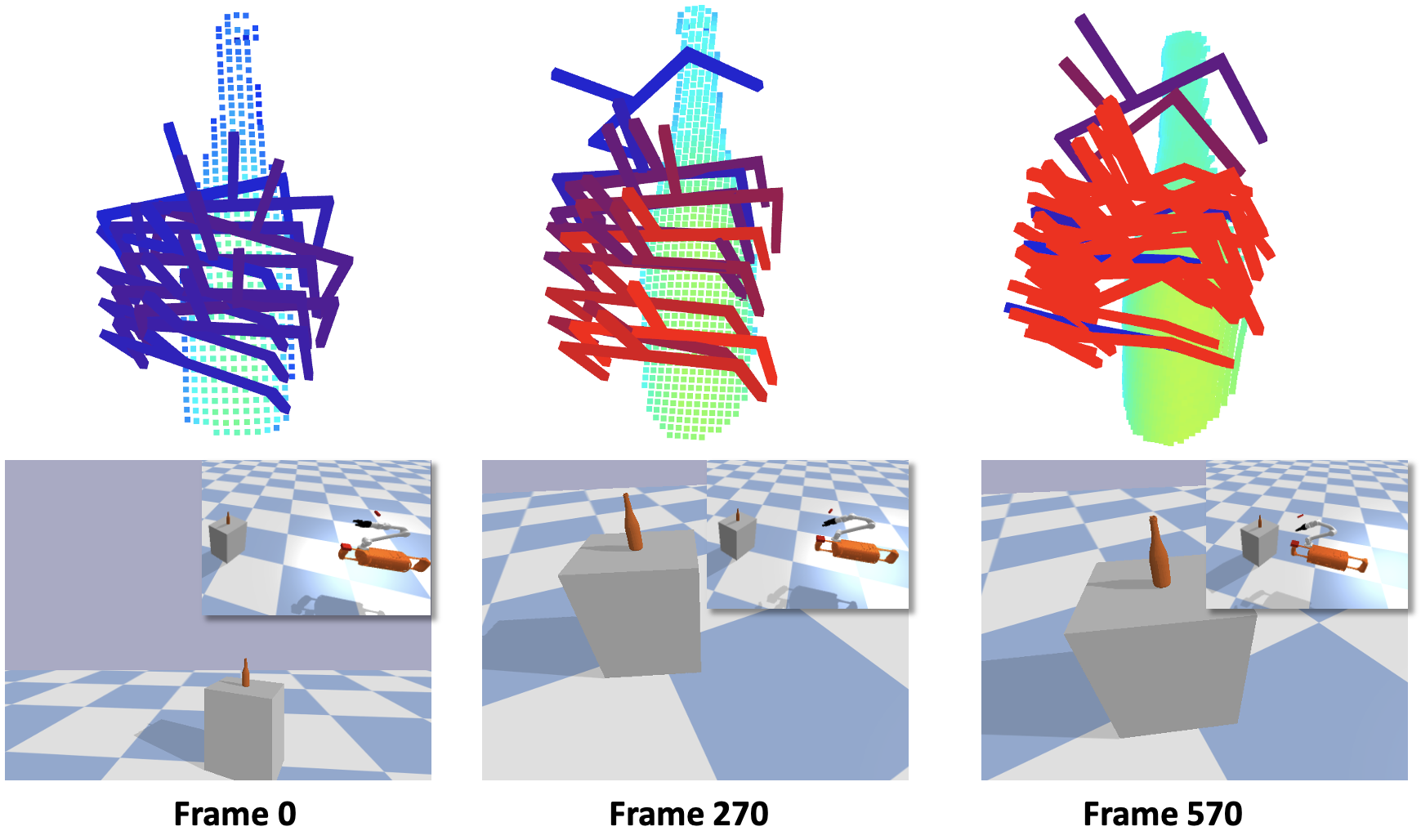}
\end{overpic}
\caption{
Visualization of the quality of the fused grasping poses during mobile manipulation. The grasping are color-coded based on their graspability feature (to red the better).
}
\vspace{-5mm}
\label{fig:policy}
\end{figure}

\subsection{Results}

\textbf{Comparsion on Habitat simulator.} To comprehensively evaluate our method, we conduct extensive experiments in the Habitat simulator. 
These results are demonstrated in Table~\ref{tab:benchmark}. Here, we find that our method achieves state-of-the-art performance in all challenging settings.
%
Moreover, we find other methods have an apparent performance drop from non-cluttered episodes to cluttered episodes because the cluttered scenes required more accurate gripper poses to avoid mistakingly grasping other objects. Our methods benefit from temporally consistent grasping poses and directly learn how to drive the gripper to the grasping pose, showing only a small performance drop. 
Another interesting finding is that the GazeSR and GraspSR have large performance gap in many methods that use abstract grasp (M3~\cite{Gu2022MultiskillMM}, HB~\cite{Szot2021Habitat2T}, ReachMM), which proved that the grasping poses perform better to motivate grasping strategy.

\textbf{Comparsion in Isaac Gym and real-world environments.}
We deploy our policy, trained on the Isaac Gym simulator, to real-world environments (Unitree B1 + Z1 arm). The results on both Isaac Gym and the real-world environment are reported in Table~\ref{tab:comp_real}. We evaluate two widely-demanded skills grasping within cluttered objects and avoiding obstacles simultaneously. 
Our findings indicate that directly deploying our method yields robust performance in real-world settings. Furthermore, when comparing with the GAMMA without fusion, we observed a performance drop from GazeSR to GraspSR, a trend also identified in the Habitat~\cite{Szot2021Habitat2T} environment(Table~\ref{tab:benchmark}). This supports the importance of complete and accurate graspability in RL training. 

%


\textbf{Mobile manipulation process analysis.} 
We plot two main observations related to graspability-aware mobile manipulation: the number of grasping poses and the distance from the gripper to these poses (Equ.~\ref{equ:dgg}). The results, showcased in Fig.\ref{fig:policy}, originate from the Habitat environment. These data compellingly indicate that our policy adeptly directs the agent to identify an increased number of grasping poses while simultaneously nearing those specified positions effectively. Additionally, to elucidate the significance of the graspability state, we color-code the integrated grasping pose and employ a fine voxel size of $1$cm to ensure a detailed visualization. Throughout the mobile manipulation process, it's evident that the agent gains confidence while observing and approaching the grasping poses.

\begin{figure}[t]
\centering
\begin{overpic}
[width=0.93\linewidth]
{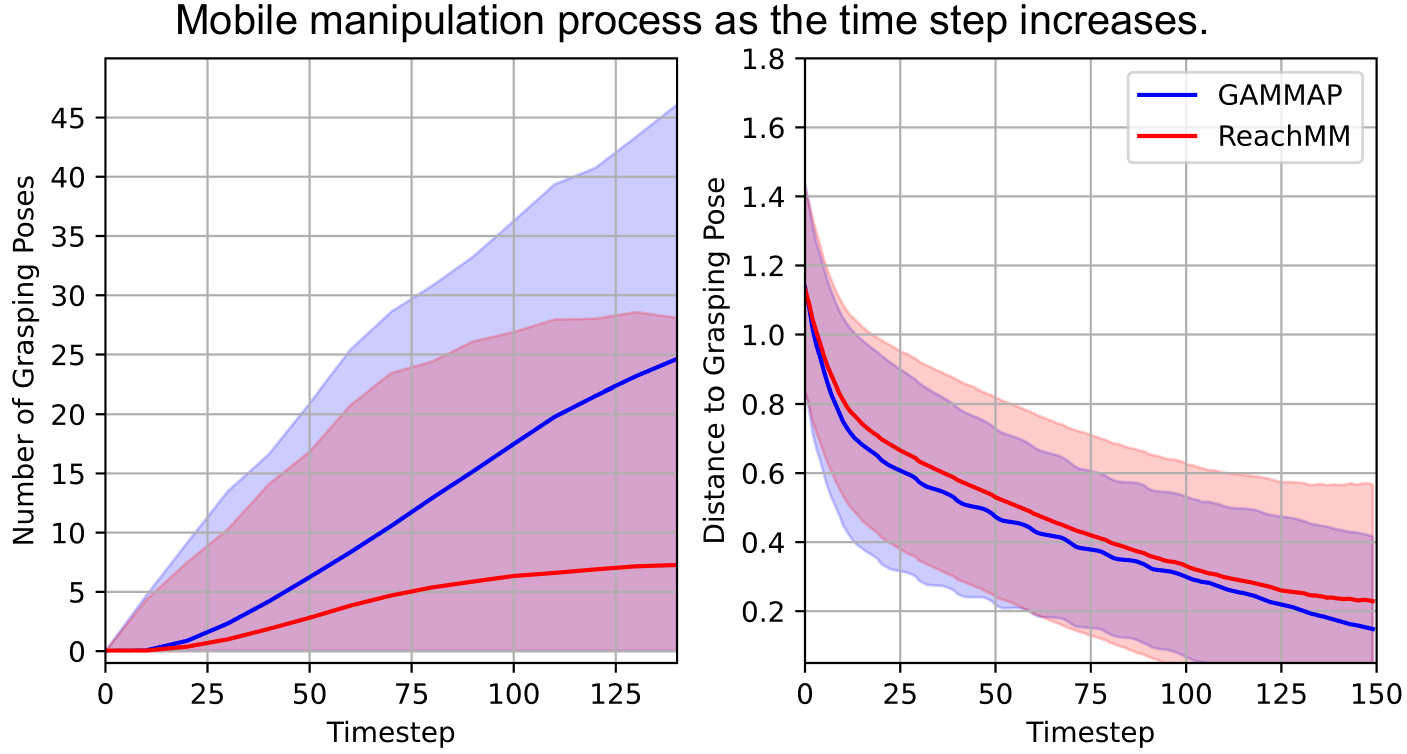}
\end{overpic}
\caption{
Compared with ReachMM in terms of (1) the number of grasping poses and (2) the distance to the grasping poses.
}
\vspace{-5mm}
\label{fig:policy}
\end{figure}

\textbf{Runtime and memory analysis.}
The online grasping fusion module is highly efficient, both in terms of memory usage and execution speed. It demands less than 0.1 GB of memory for a single scene and operates at real-time frame rates. For training, our method necessitates 36 GPU hours on an A100 to attain state-of-the-art performance.

\section{CONCLUSIONS}


We introduce a graspability-aware mobile manipulation policy, enabling agents to achieve robust and accurate mobile grasping. This capability is powered by an online grasping pose fusion module, which fuses online predicted grasping poses. This leads to temporally consistent grasping pose observations, facilitating learning graspability. Our method demonstrates superior performance on Habitat. We also deploy our approach on the Unitree B1 with Z1 arm for real-world experiments, further showcasing the robustness of our methodology. In the future, we would like to explore the potential of this framework in highly dynamic environments or for long-horizon tasks.

\section{Acknowledgment}
We thank all reviewers for their insightful comments and valuable suggestions.  This project is supported by the National Natural Science Foundation of China (No. 62306016) and Beijing Academy of Artificial Intelligence (BAAI).










\bibliographystyle{IEEEtran}
\bibliography{IEEEabrv,reference}

\end{document}